# Progressive extension of reinforcement learning action dimension for asymmetric assembly tasks

Yuhang Gai
*Department of Mechanical Engineering*
*Tsinghua University*
Beijing, China
gaiyh14@126.com

Jiuming Guo
*Department of Mechanical Engineering*
*Tsinghua University*
Beijing, China
guojm-thu@qq.com

Dan Wu
*Department of Mechanical Engineering*
*Tsinghua University*
Beijing, China
wud@tsinghua.edu.cn

Ken Chen
*Department of Mechanical Engineering*
*Tsinghua University*
Beijing, China
kenchen@tsinghua.edu.cn

*Abstract*—**Reinforcement learning (RL) is always the preferred embodiment to construct the control strategy of complex tasks, like asymmetric assembly tasks. However, the convergence speed of reinforcement learning severely restricts its practical application. In this paper, the convergence is first accelerated by combining RL and compliance control. Then a completely innovative progressive extension of action dimension (PEAD) mechanism is proposed to optimize the convergence of RL algorithms. The PEAD method is verified in DDPG and PPO. The results demonstrate the PEAD method will enhance the data-efficiency and time-efficiency of RL algorithms as well as increase the stable reward, which provides more potential for the application of RL.**

***Keywords—reinforcement learning, efficient exploration, robot force control, intelligent control***

## I. Introduction

Robot automatic assembly is one of the mainstream directions of industrial development. The method of force control is one of the main technical paths of robot automatic assembly. The core of the force control method is to construct the mapping between the operating force and the relative pose of the assembly objects. The mapping of symmetric objects can be easily modeled [1] [2]. Then some model-based methods, such as compliance control [3] [4], can be used to control the robot to implement the assembly task. However, the mapping of the asymmetric objects is usually unattainable and polysemous through modeling. Therefore, model-free RL methods will be a preferred embodiment to construct the control strategy of asymmetric assembly tasks.

The dilemma is that the RL methods always have a local attribute. Once the environment or task changes, the learned agent will be inappropriate and needs to be retrained. Therefore, the convergence speed of RL becomes the crux of practical application.

The motivation of this paper is to propose a method to construct the strategy for symmetric assembly tasks efficiently. Specifically, the original intention includes not only the construction of the control strategy but also the improvement of the convergence speed.

The strategies to improve the convergence speed are summarized as follows: parallel training [5] [6] [7], exploration guidance [8] - [14], and exploitation optimization [15] [16].

Some RL algorithms open up multiple threads for parallel training, typical of which are A3C [5] and DPPO [6]. Besides, [7] further proposed to accelerate convergence through parallel training of multiple GPUs.

Exploration guidance strategies can usually improve the convergence independent of RL algorithms. The most basic strategy is to guide the exploration of RL through a basic model [8] - [12]. The RL agent assists the model-based compliance control to complete the assembly tasks. Adaptive exploration is another strategy to accelerate convergence. Plappert [13] proposed the method of adaptive exploration noise to balance exploration and convergence. Discretizing the action space is also an effective method to accelerate the convergence [14].

Exploitation optimization strategies are always related to RL algorithms. For example, off-policy RL algorithms always tend to converge much faster than on-policy RL algorithms because of the experience replay mechanism. Therefore, off-policy RL algorithms, such as DDPG [15] and DQN [16], seem to be more common in the field of industrial control. PPO [17] employs the importance sampling mechanism, which also accelerates the convergence.

The main contributions of this paper are as follows. First, we combine the RL algorithms and compliance control to conduct asymmetric assembly tasks, which achieves better performance and accelerates convergence. Another more important contribution of this paper is to propose a completely innovative PEAD mechanism to accelerate the convergence of RL algorithms, which can work in conjunction with the above methods together as well as independently. To implement the PEAD method, a mapping between the high-dimensional action space and low-dimensional action space is established. Based on the mapping, the corresponding general network extension method is proposed. The PEAD method is verified in DDPG and PPO algorithms.

The rest of the paper is organized as follows. Section II introduces the method of constructing control strategies through combining RL and compliance control. Section III develops the PEAD mechanism to accelerate the convergence of RL algorithms. The simulation and the experiment are provided in Section IV. Section V summarizes the research work of this paper.

## II. Control Strategy

In this section, we first formulate the problem. Then we introduce the solution framework to construct the control strategy through combining a RL agent and a compliance controller.

### *A. Problem Formulation*

This paper focused on settling asymmetric assembly tasks through the method of force control. The situation is shown in Fig. 1. The assembly objects are a multi-peg group and a multi-hole group. The multi-hole group is fixed on the table. A position-based robot manipulates the multi-peg group to complete the assembly task in Cartesian space continuously. A force-moment sensor is fixed at the end of the robot, which perceives the operating forces and moments.

The assembly task can be formulated as a Markov decision process (MDP) [18], then RL algorithms can be applied to construct the control strategy. This paper combines the residual RL in [12] and the compliance control in [3] to complete the assembly task.

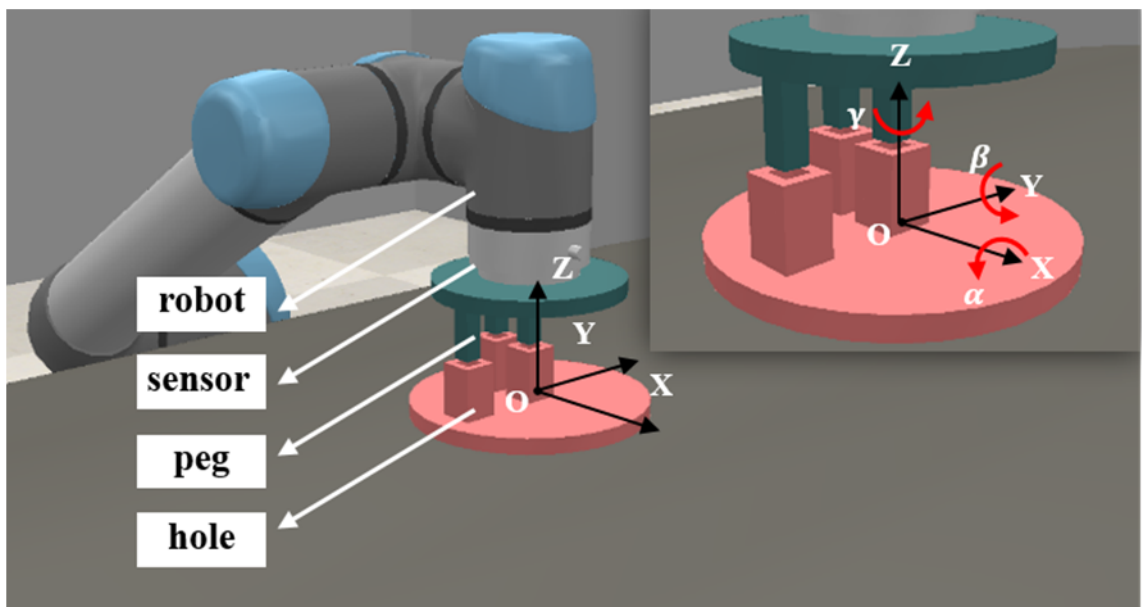

Fig. 1. The asymmetric assembly task and the manipulation of the robot.

A 12-dimensional state space $\mathcal{S}$ is constructed by integrating the Cartesian pose of the robot with the six-dimensional sensor information. A state $s_t \in \mathcal{S}$ at time $t$ is

$$s_t = \left[x, y, z, \alpha, \beta, \gamma, F_x, F_y, F_z, M_x, M_y, M_z\right] \tag{1}$$

where the Cartesian pose of the robot is defined in Fig. 1, while the forces and moments are parallel with the corresponding pose elements.

The actions of RL are used to correct the translations and rotations of the robot commanded by the compliance controller. A six-dimensional action space $\mathcal{A}$ is spanned by the correction factors. An action $a_t \in \mathcal{A}$ at time $t$ is

$$a_t = \left[\Delta x, \Delta y, \Delta z, \Delta\alpha, \Delta\beta, \Delta\gamma\right] \tag{2}$$

To execute the assembly task, an agent is needed to construct a policy $\pi : \mathcal{S} \mapsto \mathcal{A}$, which can map the state space to the action space. The policy $\pi$ will decide the next action according to the current state.

$$a_{t+1} = \pi(s_t) \tag{3}$$

The assembly process is always expected to be as quick as possible, while the operating forces and torques are required to be limited enough. These targets will be defined through a reward function $r(s_t, a_t, s_{t+1})$. The long-term value can be evaluated through the sum of discounted rewards

$$Q_\pi(s,a) = \mathbb{E}^\pi\left[\sum_{i=0}^{\infty} {\gamma_{dis}}^i r(s_{t+i}, a_{t+i}, s_{t+i+1}) \mid s_{t+i} \in \mathcal{S}, a_{t+i} \in \mathcal{A}\right] \tag{4}$$

where $\gamma_{dis} \in [0,1]$ is the discount rate. $\mathbb{E}^\pi[\cdot]$ is the expectation function when the state-action pairs follow the policy $\pi$.

The terminal goal of the assembly task is to train an optimal policy $\pi^*$ to maximize the value function.

$$\pi^*(s) \in \arg\max_{a \in \mathcal{A}} Q_\pi(s,a) \tag{5}$$

### *B. Solution Framework*

The framework is illustrated in Fig. 2. The robot is driven by a planner output $p_p$ and a controller output $p_c$. The planner plans the trajectory of an assembly task offline. Here we just

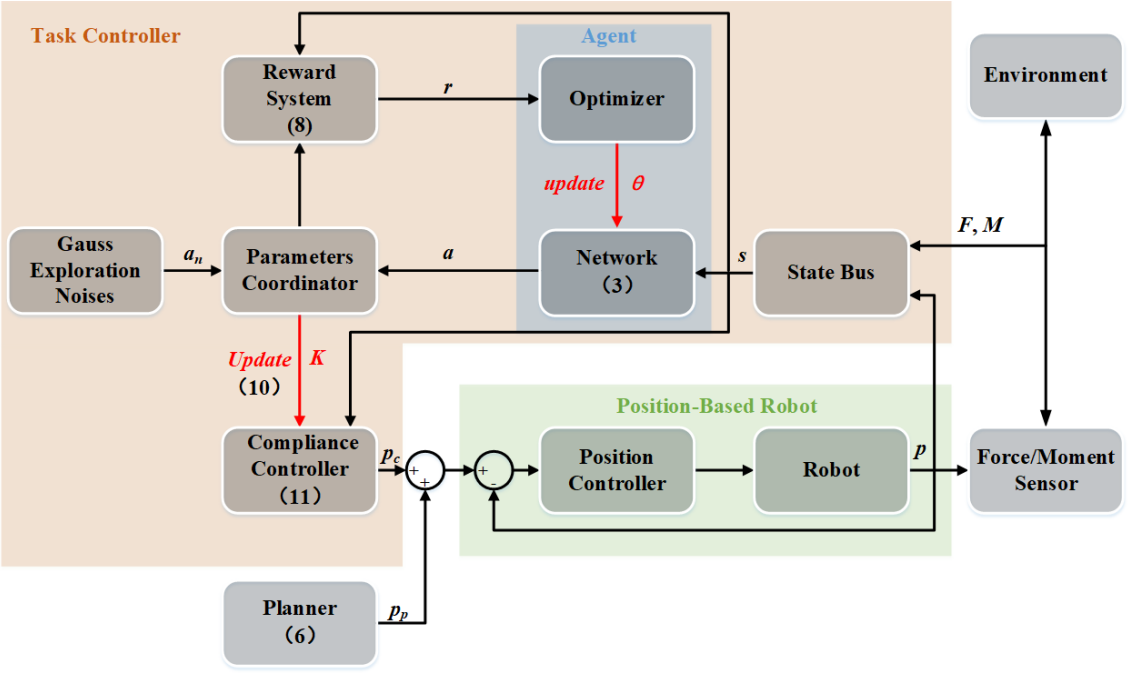

Fig. 2. The control framework of combining RL and compliance control.

plan a trajectory with a constant velocity. The output of the planner $p_p$ is

$$p_p = \left[0, 0, L/step, 0, 0, 0\right] \tag{6}$$

where $L$ indicates the target insertion depth. *step* indicates the desired assembly step.

Apart from the offline planner, the basic framework is completing a force close-loop outside the position-based root. The task controller is composed of a compliance controller and a residual RL agent. The agent is used to update the compliance controller. The compliance controller is selected as a proportional controller and utilizes the format in [3].

$$\hat{p}_c = \hat{\mathbf{K}}\mathbf{A}(F - F_{rfr}) \tag{7}$$

where $\hat{\mathbf{K}} = \mathrm{diag}(k_x, k_y, k_z, k_\alpha, k_\beta, k_\gamma)$ is the diagonal matrix of proportional parameters, whose elements are both constant. $\mathbf{A}$ is the direction matrix, which is used to describe the equivalent effect of forces and moments. $F_{rfr}$ represents the reference force during the assembly. When the pegs and holes are with clearance fit, all the elements of $F_{rfr}$ are zeros.

The reward function guides the agent to reduce assembly time and assembly stress. The reward function is defined as

$$r = r_d + r_s \tag{8}$$

where $r_d$ is the dense reward. The dense reward is

$$r_d = -h_z p_z - h_F \left\|F\right\|_2 - h_M \left\|M\right\|_2 \tag{9}$$

where $h_z$, $h_F$, $h_M$ indicate the coefficients designed manually. $p_z$ indicates the motion of the robot along the Z direction. $F = [F_x, F_y, F_z]$ and $M = [M_x, M_y, M_z]$ represent the force vector and moment vector. As $p_z$ is desired to be negative, the first term is used to encourage the insertion. The last two terms are used to punish the excessive operation forces and moments.

$r_s$ is the sparse reward to evaluate a training episode. If the pegs can be inserted to target depth $L$ and forces and moments of the entire assembly process are all within the limitation, the total episode will be rewarded by $r_s = 1$. On the contrary, if the forces or moments are overloaded, the assembly is failed and will be punished by $r_s = -1$.

The agent is set to output a correction factor $a_{rl} \in \mathbb{R}^{1\times 6}$. All the elements of the action $a$ are set within the range of $[lb, ub]$. For some RL algorithms, exploration noises are needed. Here we generate Gauss noises $a_n \sim \mathrm{N}(\mu, \sigma^2 \mathrm{I}_6)$ to extend the exploration of the agent. In this paper, only action noises are implemented and parameter noises are omitted. Both the action of the agent and the Gauss noises are used to update the compliance parameters.

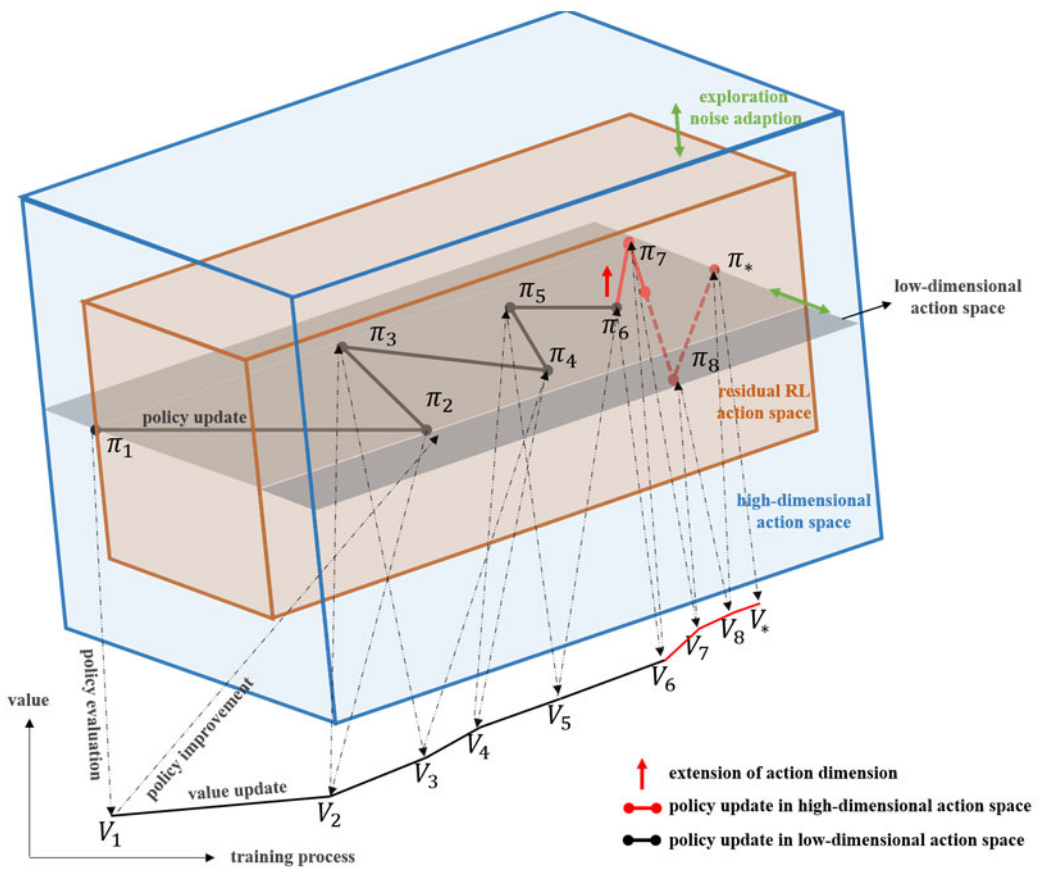

Fig. 3. The relation between the exploration in action space and the convergence of RL.

$$\mathbf{K} = \mathrm{diag}((a + a_n)\hat{\mathbf{K}} + \hat{\mathbf{K}}) \tag{10}$$

Then the actual command pose $p_c$ becomes

$$p_c = \mathbf{KA}(F - F_{rfr}) \tag{11}$$

DDPG is a typical off-policy RL algorithm. The actor-critic structure and experience replay mechanism guarantee the algorithm is quite efficient. PPO is a typical on-policy RL algorithm. The sampling method makes PPO algorithm stable enough. Therefore, we select these two RL algorithms to train the agent separately. Besides, it is persuasive to verify the generality of the PEAD method on different algorithms.

## III. Progressive Extension of Action Dimension

In this section, we first analyze how the exploration process in the action space affects the convergence of RL. Then we propose the PEAD method and apply the PEAD method to two different RL algorithms.

### A. Exploration in Action Space

Optimality and convergence speed are contradictory in some aspects. Optimality requires that the exploration of RL algorithms should be large enough to get rid of local optimums. But the convergence speed requires that the exploration should be limited. Therefore, proper exploration is crucial for RL algorithms.

The training process of RL is abstracted as exploring and exploiting a sequence of policies and values in the policy space and the value space. During each iteration of the training process, the policy will go through a policy evaluation and a policy improvement [19]. After a complete iteration, the updated policy tends to get closer to the optimal strategy, while the value tends to increase.

An intuitive assumption is that the volume of the policy space is related to that of the action space. An agent usually needs to learn an extremely complex and completely unfamiliar knowledge. If the action space is too large, the convergence speed and stability of the training process cannot be guaranteed. Hence, it is meaningful to optimize the exploration in the action space.

Residual RL combines basic knowledge to achieve the target of optimizing exploration. Taking the assembly task as an example, residual RL can constrain exploration in a smaller action space around the basic compliance model. Exploration noise adaption shares the same idea, which constrains exploration within manually designed limits to balance the convergence speed and exploration.

The PEAD method proposed in this paper optimizes exploration from a completely different perspective. We first construct a mapping from the high-dimensional action space to the low-dimensional action space. Then we approximate the training in the high-dimensional action space to that in the low-dimensional one. As shown in Fig. 3, the exploration in the low-dimensional action space will be more efficient. However, the optimality in the low-dimensional action space is not as good as that in the high-dimensional one. Therefore, the extension of action space is needed at a proper time. Then a subsequent training can be executed to achieve a higher reward.

### B. PEAD Method

The basic assumption of the PEAD method is that there is a similarity or correlation between actions, which can be approximated into each other through certain transformations. For symmetric objects like cylinders, the spatial control strategy can be decomposed into two orthogonal plane control strategies [3]. These two plane control strategies are identical, thus making the policies in two planes the same. Therefore, the low-dimensional action space is actually equivalent to a high-dimensional one. For asymmetric assembly tasks, the low-dimensional action space and the high-dimensional action space are not equivalent, but the similarity still exists, which is fundamental for the PEAD method.

We define three action spaces to demonstrate how to apply the PEAD method. The action dimension of three action spaces is from low to high.

$$\begin{aligned} {}^l\pi(s_t) &= \left[{}^l\pi_1(s_t), {}^l\pi_2(s_t), \ldots, {}^l\pi_l(s_t)\right] \\ {}^m\pi(s_t) &= \left[{}^m\pi_1(s_t), {}^m\pi_2(s_t), \ldots, {}^m\pi_m(s_t)\right], l < m \\ {}^n\pi(s_t) &= \left[{}^n\pi_1(s_t), {}^n\pi_2(s_t), \ldots, {}^n\pi_n(s_t)\right], m < n \end{aligned} \tag{12}$$

The extension will first occur from ${}^l\pi(s_t)$ to ${}^m\pi(s_t)$, later from ${}^m\pi(s_t)$ to ${}^n\pi(s_t)$. The extension can continue until the action dimension can realize the target of the task. Here we just explore the extension from ${}^l\pi(s_t)$ to ${}^m\pi(s_t)$. The PEAD method first constructs a mapping from high-dimensional action space to low dimensional action space.

$${}^l\pi(s_t) = \mathrm{f}({}^m\pi(s_t)) \tag{13}$$

Linearize the mapping to obtain a similar transformation.

$${}^l\pi(s_t) = {}^m\pi(s_t)\mathbf{F}, \mathbf{F} \in \mathbb{R}^{m\times l} \tag{14}$$

The function f and the matrix F can be used to describe the similarity or correlation between actions and then reduce the dimension of action space. To obtain a higher reward, it is necessary to extend the action space dimension at a proper time, the extension can be achieved through the inverse mapping. The inverse mapping and its linearization are

$$\begin{aligned} {}^m\pi(s_t) &= \mathrm{f}^{-1}({}^l\pi(s_t)) \\ {}^m\pi(s_t) &= {}^l\pi(s_t)\,\mathrm{pinv}(\mathbf{F}), \mathbf{F} \in \mathbb{R}^{m\times l} \end{aligned} \tag{15}$$

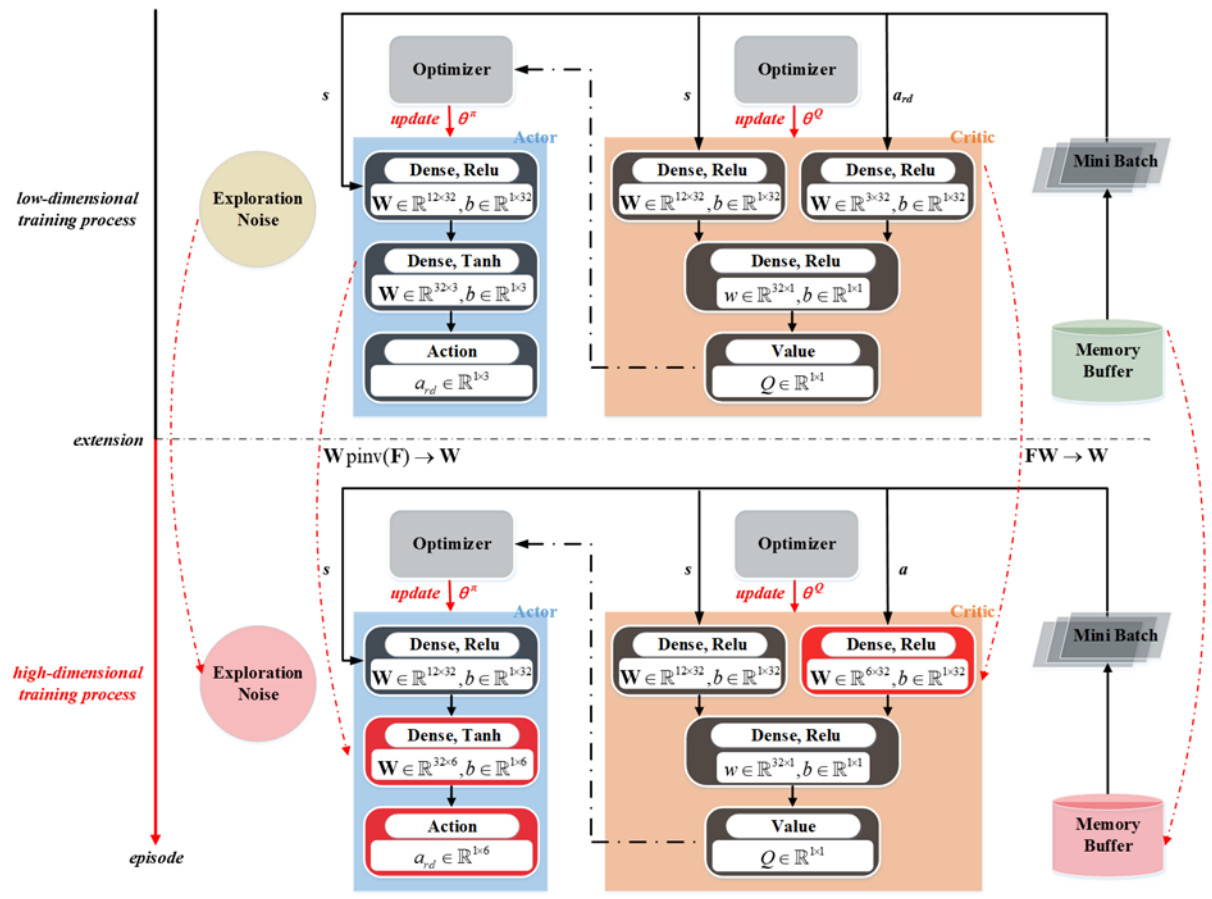

Fig. 4. Implementation of the PEAD method on DDPG algorithm. The target networks are not shown.

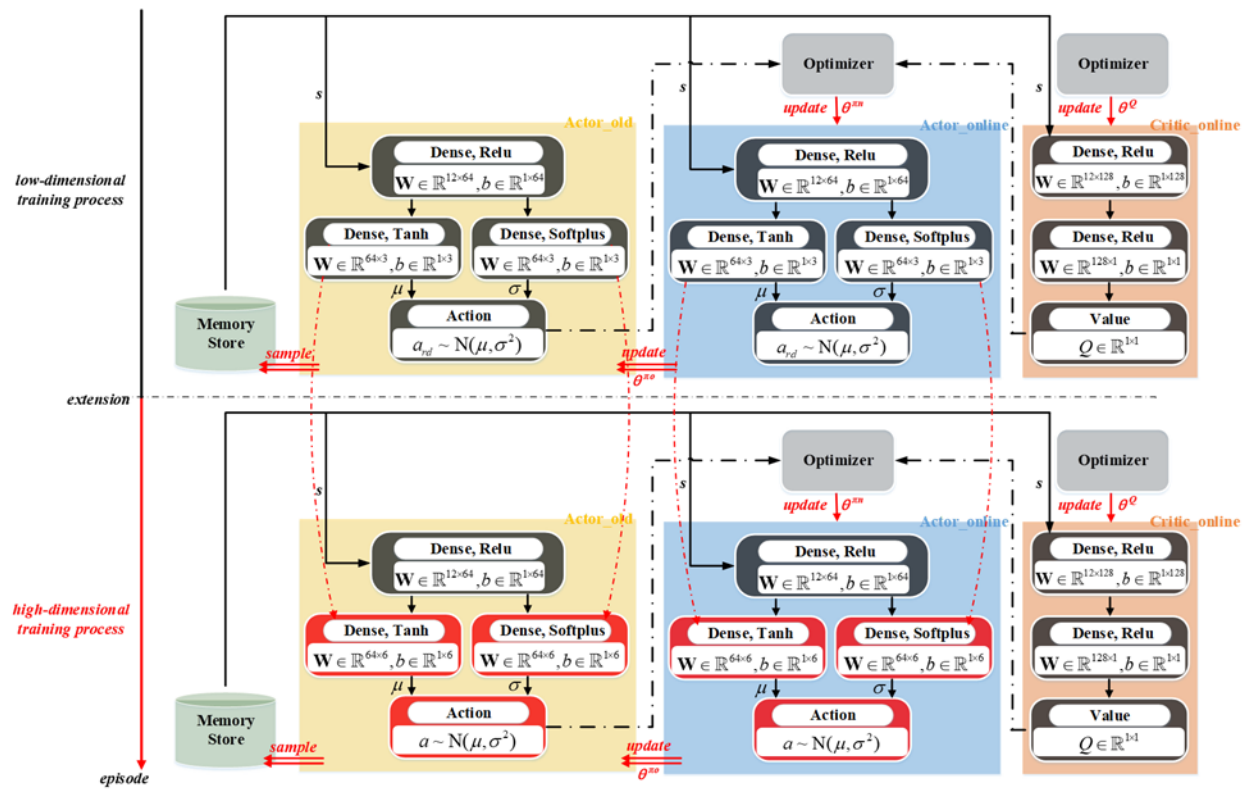

Fig. 5. Implementation of the PEAD method on PPO algorithm.

The extension can be implemented repeatedly to progressively extend the action dimension.

### C. Implementation of the PEAD method

In this subsection, we explain how to implement the PEAD method to adjust the policy parameters of a RL agent. The components of action $a$ defined in (2) are used to modify the compliance parameters $k_x$, $k_y$, $k_z$ $k_\alpha$, $k_\beta$, $k_\gamma$ one by one. It is reasonable to assume that $\Delta x$ and $\Delta y$ are in a similar policy pattern. While $\Delta\alpha$, $\Delta\beta$, and $\Delta\gamma$are in another policy pattern. Therefore, the mapping matrix F is

$$\mathbf{F}=\begin{bmatrix}1/2 & 1/2 & 0 & 0 & 0 & 0\\ 0 & 0 & 1 & 0 & 0 & 0\\ 0 & 0 & 0 & 1/3 & 1/3 & 1/3\end{bmatrix}^{\mathrm{T}} \tag{16}$$

The dimension-reduced action $a_{rd}$ is

$$a_{rd}=\left[\Delta p,\Delta z,\Delta o\right] \tag{17}$$

The dimension-reduced action $a_{rd}$ will replace the action $a$ to modify all the compliance parameters. Specifically, action $\Delta p$ will modify $k_x$, $k_y$. Action $\Delta z$ will modify $k_z$. Action $\Delta o$ will modify the $k_\alpha$, $k_\beta$, $k_\gamma$ in the same manner as (10).

When implementing the PEAD method, the RL agent needs to be adjusted accordingly. We exhibit the application of the PEAD method on DDPG, a typical off-policy RL algorithm, and PPO, a typical on-policy RL algorithm.

#### 1) DDPG

DDPG algorithm works within the actor-critic framework [15]. The actor is an online policy network used to supply action for the current state. The critic is an online value network to evaluate the value of the action-state pair. After a policy evaluation in the online value network, the optimizer updates the value network and the policy network in turn.

To offer the target values for updating, the online policy network and the online value network are copied as a target policy network and a target value network. To guarantee convergence and stability, the parameters of the target policy network and the target value network are soft updated slowly.

When implementing the PEAD method, the training process is divided into three stages: training in low-dimensional action space, network extension, and training in high-dimensional action space. The extension scale is defined as the number of episodes when extending the network. The entire PEAD method applied to DDPG is illustrated in Fig. 4.

The actor and critic network of DDPG can be defined as network parameters $\theta^{\pi}$ and $\theta^{Q}$. Each dense layer can be described as a weight matrix $\mathbf{W}$ and a bias $b$.

During the low-dimensional training process, the input of the critic and the output of the actor are dimension-reduced action $a_{rd}$. The shape of the weight matrix of the input dense layer of the critic is $3\times32$ and that of the output dense layer of the actor is $32\times3$. After extension, the dimension-reduced action $a_{rd}$ will be replaced by high-dimensional action $a$. The shapes above will become $6\times32$ and $32\times6$.

The basic criteria for extension are as follows. First, the actions supplied by the actors for the same state before and after the extension follow the equality constraints of (13), (14), and (15). Second, the values predicted by the critic for the same action-state pair before and after the extension are equal. As indicated by (13), (14), and (15), the weight matrix of the output dense layer of the actor is updated through

$$\mathbf{W}\,\mathrm{pinv}(\mathbf{F})\rightarrow\mathbf{W} \tag{18}$$

The weight matrix of the input dense layer of the critic is updated through

$$\mathbf{FW}\rightarrow\mathbf{W} \tag{19}$$

The updates are carried out simultaneously on both the online networks and the target networks. The exploration noise and memory buffer also need to be updated in a similar manner to (15).

#### 2) PPO

PPO also works within the actor-critic framework. Different from DDPG, PPO is an on-policy RL algorithm. Hence, PPO has only online networks and no target networks.

To improve data utilization, the importance sampling mechanism is applied. Hence, a copy of the online actor is designed, which is called the old actor. The old actor is updated at a lower frequency than the online actor and the online critic. After the online networks have been updated several times, the old actor is regularly updated by the online actor. Then the memory store is cleared and resampled. When the memory store is filled, the training of online networks continues.

The extension scale is selected just after the old actor is updated and before the memory store is sampled. The extension process is illustrated in Fig. 5.

The old actor and online networks are parameterized as $\theta^{\pi o}$, $\theta^{\pi n}$ and $\theta^{Q}$. Each dense layer is also described as a weight matrix **W** and a bias $b$.

As the input of the online critic is only state, the extension of action dimension has no influence on the critic. The actor consists of two dense layers, which represent the mean and variance of the output. The output layer is a normal distribution layer constituted by the mean and variance dense layers. During the low-dimensional training process, the shape of the mean and variance dense layers of the actor is $64\times3$. After extension, the dimension-reduced action $a_{rd}$ will be replaced by high-dimensional action $a$. The shape above will become $64\times6$. The transformation mapping also satisfies (18).

Different from DDPG, PPO employs the stochastic policy. Therefore, PPO does not need noises to explore the action space. Besides, the memory store is cleared and resampled periodically. Because we extend action dimension just after the old actor updates and the memory store will be resampled after extension, there is no need to extend the memory store.

## IV. Simulation and Experiment

### A. VREP Setup

The verification is implemented in CoppeliaSim software and driven by physical engine Bullet Physics. The situation is shown in Fig. 1. The assembly objects are an asymmetric part constituted by three cuboid pegs and an asymmetric part constituted by three cuboid holes.

The axis lines of the pegs and the holes can be completely coincident. The distance of pegs in the X direction is 50 mm while that in the Y direction is 40 mm. The length of pegs and holes is 30 mm and the final insertion depth is also set to be 30 mm. The target insertion step is 50 and the maximum step is 100. The cross section of the pegs and holes is all square. The side length of the square cross section of pegs is 9.9 mm while that of holes is 10 mm.

The assembly task is divided into two stages: search and insertion. The proposed control strategy focuses on the insertion stage. The search stage is accomplished by artificial teaching. The pegs are first demonstrated right above the holes accurately and the bottom of the pegs is kept at a small distance from the top of the holes. Then the search stage is recognized to be accomplished.

Then we train the force control strategy to implement the insertion stage. At the beginning of each episode, the robot is first commanded to move to the final demonstration pose of the search stage. Then the robot is commanded to move along the negative Z direction to a small insertion depth. As the search stage is accomplished by artificial teaching accurately, the small insertion depth will not cause jamming. To improve the robustness to the original aligning error, the pose of the pegs is selected randomly in a limited range, and the pose should ensure that the contact forces can be detected. Then the training process starts. The range of random pose components is set as [±0.2 mm, ±0.2 mm, ±0.2 mm, ±0.5°, ±0.5°, ±0.5°].

The constant compliance parameter is first tuned as diag(8e-3, 8e-3, 8e-5, 2e-3, 2e-3, 2e-3) and boundaries of action components are selected as [-0.8, 0.8].

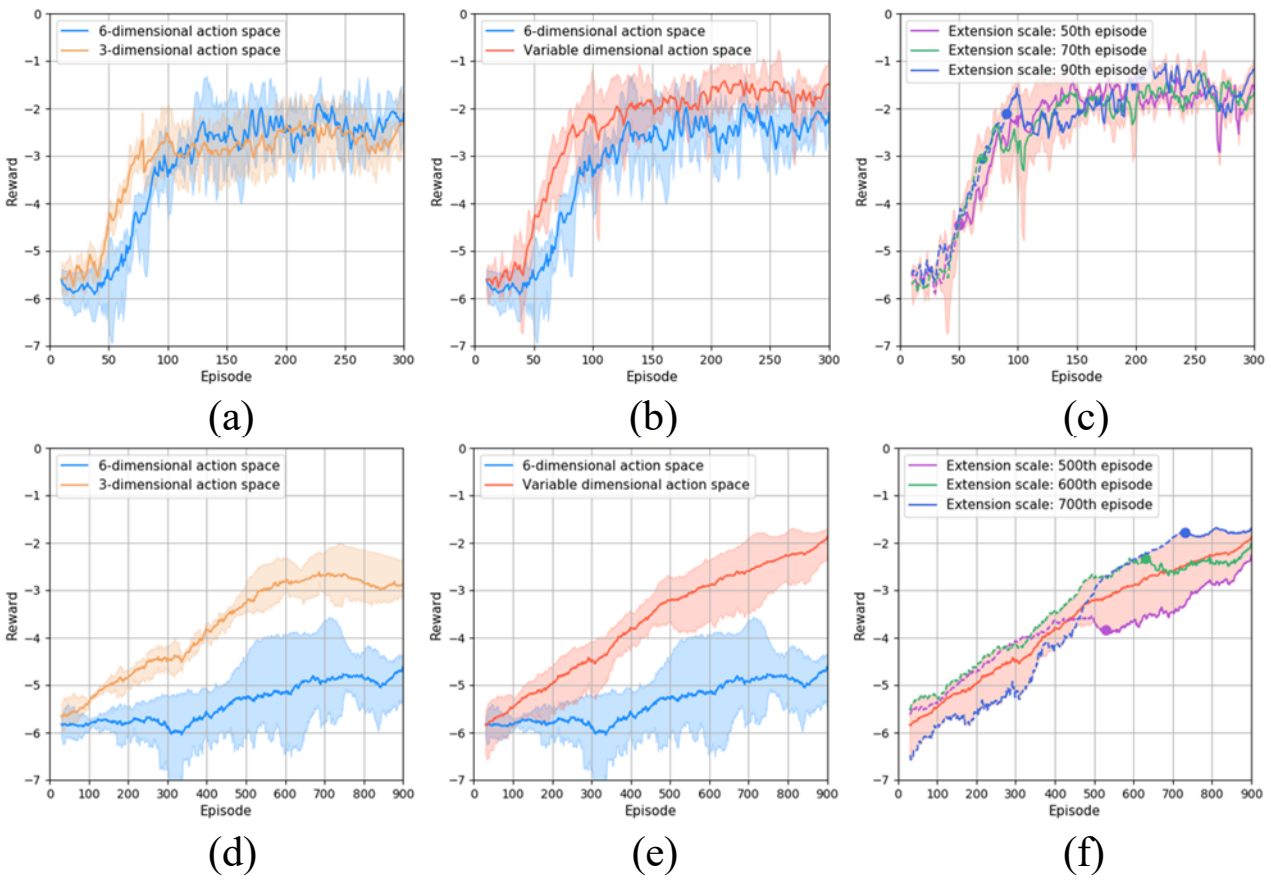

Fig. 6. The performance of DDPG and PPO during the training process. (a) - (c): DDPG. (d) - (f): PPO. (a), (d): Comparison of reward trajectories in 6-dimensional and 3-dimensional action space. (b), (e): Comparison of reward trajectories in 6-dimensional and variable dimensional action space when implementing the PEAD method. (f), (g): The performance of different extension scale.

### B. Training Process

To verify the effectiveness of the PEAD method, the training process is implemented in 3-dimensional action space, 6-dimensional action space, and variable dimensional action space separately. In each action space, the RL algorithm DDPG and PPO are used to train the agent. The learning rate of DDPG is set to be 1e-3 and 1e-2 for the actor and the critic separately. The soft update rate is set to be 1e-3. As PPO algorithm is quite unstable when the learning rate is too high, the learning rate of PPO is set to be 1e-4 and 1e-3 for the actor and critic. Coefficients in the reward function are set as $h_z$=0.5, $h_F$=0.1, $h_M$ =0.5. Because DDPG employs the experience replay mechanism, the update starts at the 30th episode. While the PPO is an on-policy algorithm, the update starts from the first episode.

The data-efficiency and final reward are the most critical evaluation indicators. The performance of data-efficiency and final reward of the PEAD method is illustrated in Fig. 6. Each training process of a certain RL algorithm in a certain action space is repeated 5 times. For a certain action space and RL algorithm, the mean reward trajectory is plotted as a solid curve and the area between the min reward trajectory and max reward trajectory is filled with a color block.

The RL algorithms in 3-dimensional action space show better data-efficiency but worse final reward. On the contrary, the RL algorithms in 6-dimensional action space perform exactly the opposite. The PEAD method combines the above advantages together. Both DDPG and PPO can achieve better data-efficiency and higher final reward after applying the PEAD method. The results prove that the proposed PEAD method is effective for improving the convergence of RL algorithms, which is of great significance to construct the control strategy of assembly tasks. The results also confirm that the final reward will get higher, which means that the assembly process becomes more smooth and causes less force and moment.

The extension scale is also important for the PEAD method. Here we just select three extension scales to show the influence. The proper scale should be chosen when the training in the low-dimensional training process is close to stable, but not yet completely stable. The principle and method to determine the extension scale remain to be further explored.

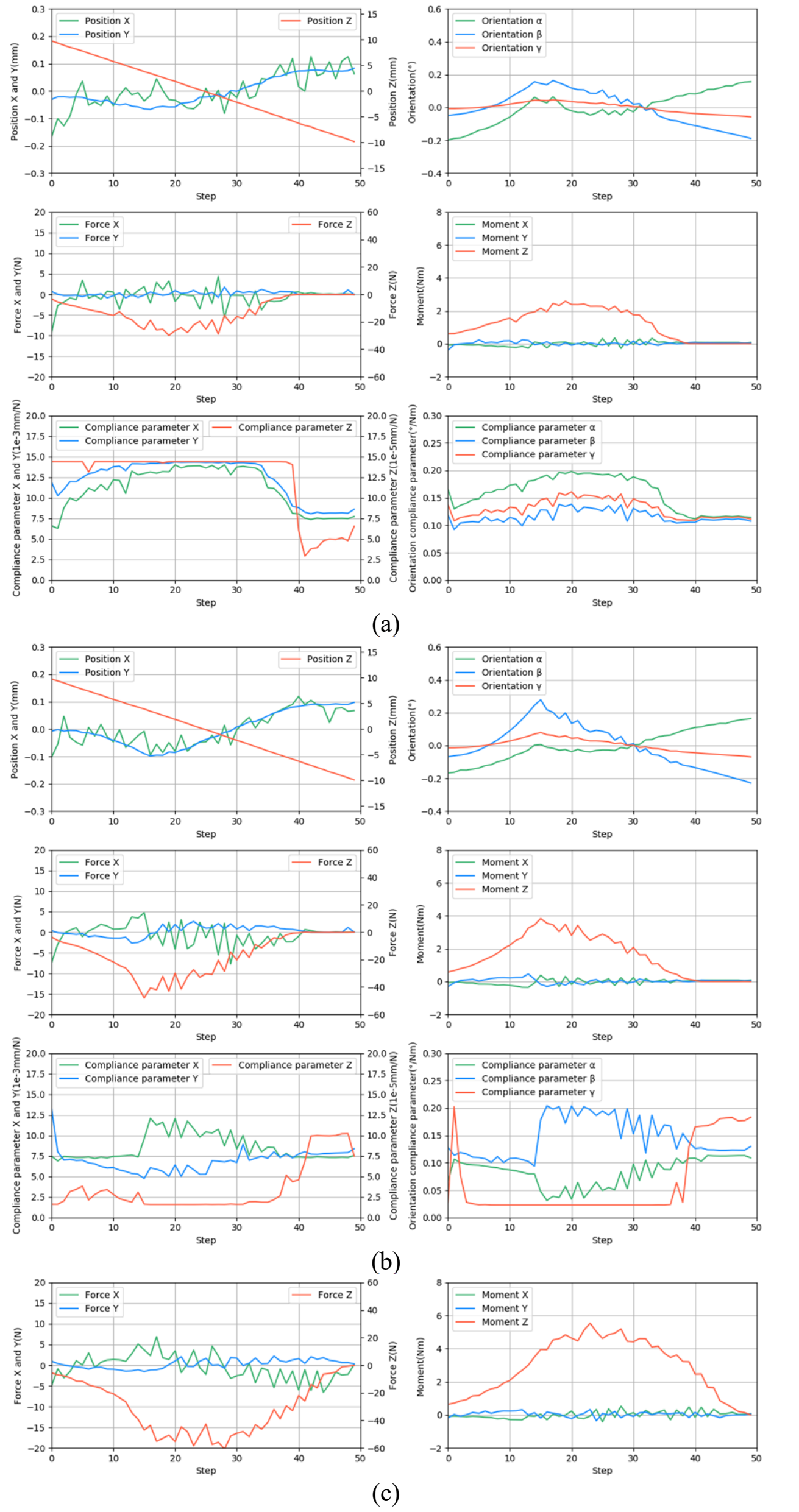

Fig. 7. The performance of control strategies. (a) Combination of DDPG and compliance control. (b) Combination of PPO and compliance control. (c) Compliance control with fixed compliance parameters. The center of position and orientation subplots are transformed to zero in (a) and (b).

The time-efficiency is also compared. Only the time spent on optimizing and extending the agent is counted, while that spent on executing assembly actions is not included. Reducing the dimensions of the action space is also of benefit in improving time-efficiency, but to a limited extent.

## C. Testing Process

We pick the agent learned by DDPG algorithm with the extension scale at the 150th episode and that by PPO at the 800th episode to construct the control strategy. The initial pose error of the robot is set as [0.2 mm, 0.2 mm, 0.2 mm, 0.5°, 0.5°, 0.5°]. Because of the interference between pegs and holes, the actual pose of the robot is not the same as the set value.

At the beginning of the assembly, the operation forces and moments continue to be larger because the interference area becomes broader. Then compliance parameters tend to close to the upper boundary to maximize the adjusting ability. Then the operation forces and moments start to decrease. With the decrease of operation forces and moments, compliance parameters also decrease to keep the assembly process more stable. The performance of PPO is not as good as DDPG, especially the skill of $\Delta\gamma$, which corresponds to the reward trajectory of the training process.

TABLE I. Comparison of Time-Efficiency

| Agent | Time-Efficiency | | |
|---|---|---|---|
| | *3-dimensional* | *6-dimensional* | *variable dimension* |
| DDPG | 628.0±18.7 s | 679.8±27.4 s | 657.9±43.6 s |
| PPO | 1563.7±55.2 s | 1625.9±52.6 s | 1596.5±78.3 s |

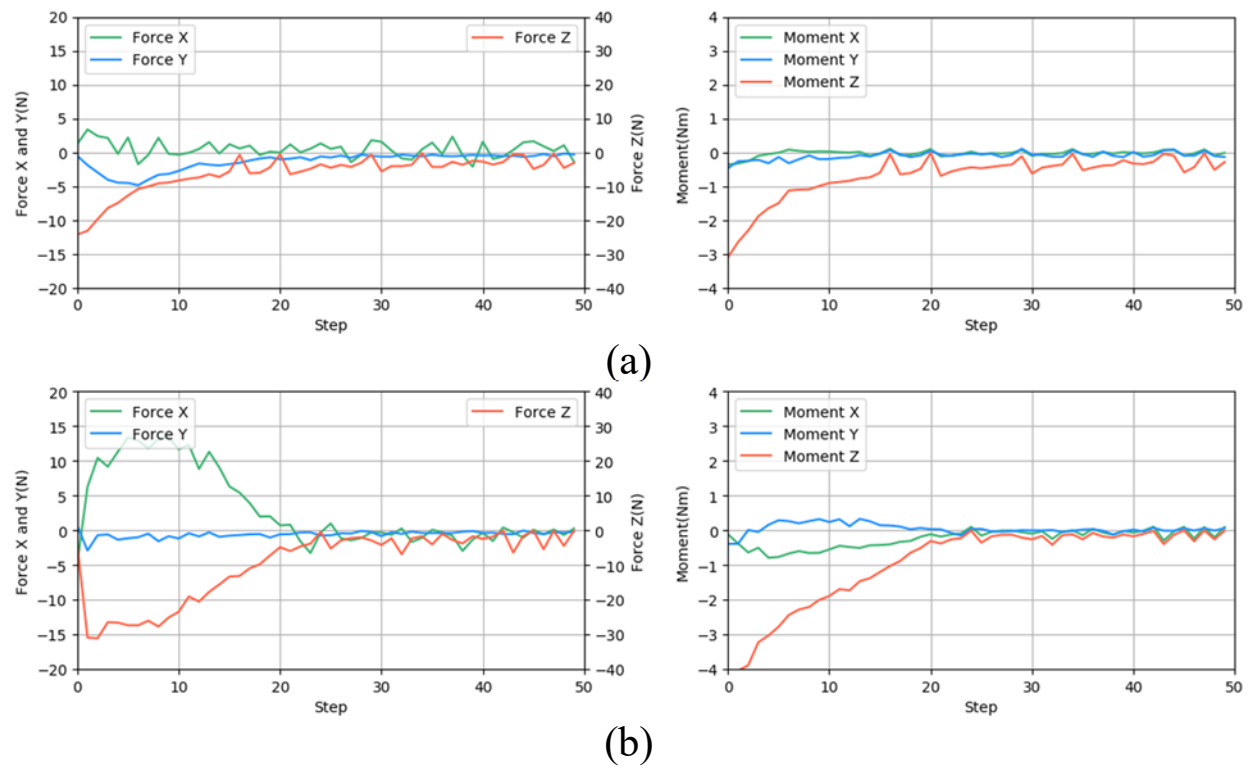

Fig. 8. The examination of the robustness of the control strategies to the initial pose error. (a) Combination of DDPG and compliance control. (b) Combination of PPO and compliance control.

Then we compare the performance of the trained control strategy with the compliance controller before training. As shown in Fig. 7, it is obvious that the trained control strategy produces less operation forces and moments, which makes the assembly much more smooth and protects the assembly objects from damage.

To test the robustness of the control strategy to the initial pose error, the initial pose error of the robot is set as [-0.2 mm, -0.2 mm, -0.2 mm, -0.5°, -0.5°, -0.5°]. As shown in Fig. 8, the control strategy can still accomplish the assembly. While the compliance controller with fixed compliance parameters cannot achieve the successful assembly. The results demonstrate that the control strategy becomes more robust after training.

## V. Conclusion

In this paper, we combine the RL algorithms with compliance control method to construct the control strategy of asymmetric assembly tasks. To accelerate the convergence of RL algorithms, we first analyze the principle of how the exploration in the action space influences the convergence of RL. Then we propose a generalized PEAD method and implement the method on DDPG algorithm and PPO algorithm separately. The results demonstrate the PEAD method can improve the data-efficiency greatly and the time- efficiency to some extent as well as can increase the stable reward of RL algorithms.

The future work is to generalize the policies obtained from existing assembly tasks to a completely new but similar task. If the generalization process is possible, the control strategies for a class of assembly objects can be transformed to each other conveniently, which will enhance the application of RL in the industry greatly.